\documentclass{IOS-Book-Article}

\usepackage{mathptmx}
\usepackage{soul}\setuldepth{article}

\usepackage[noend]{algpseudocode}
\usepackage{algorithm}
\usepackage{algorithmicx}
\algrenewcommand\textproc{}% Used to be \textsc

\usepackage{microtype}
\usepackage{multicol}
\usepackage{paralist}
\usepackage{enumitem}
\usepackage{amssymb}
\usepackage{amsfonts}
\usepackage{mathtools}
\usepackage{autoaligne}
\usepackage{multirow}
\usepackage{xspace}
\usepackage{pgf, tikz}
\usetikzlibrary{arrows, automata}

\newtheorem{example}{Example}[section]
\newtheorem{definition}{Definition}[section]

\def\hb{\hbox to 10.7 cm{}}

\begin{document}

\pagestyle{headings}
\def\thepage{}

\begin{frontmatter}              % The preamble begins here.

\title{Argument Schemes for Explainable Planning}
\author[A]{\fnms{Quratul-ain} \snm{Mahesar}}
and
\author[A,B]{\fnms{Simon} \snm{Parsons}}

% \runningauthor{Q. Mahesar et al.}
\address[A]{Department of Informatics, King's College London,
 UK}
\address[B]{School of Computing Science, University of Lincoln, Lincoln, UK}

\begin{abstract}
Artificial Intelligence (AI) is being increasingly used to develop systems that produce intelligent solutions. However, there is a major concern that whether the systems built will be trusted by humans. In order to establish trust in AI systems, there is a need for the user to understand the reasoning behind their solutions and therefore, the system should be able to explain and justify its output. In this paper, we use argumentation to provide explanations in the domain of AI planning. We present argument schemes to create arguments that explain a plan and its components; and a set of critical questions that allow interaction between the arguments and enable the user to obtain further information regarding the key elements of the plan. Finally, we present some properties of the plan arguments.

\end{abstract}

\end{frontmatter}
% \markboth{January 2020\hb}{January 2020\hb}
%\thispagestyle{empty}
%\pagestyle{empty}

\section{Introduction}
Artiﬁcial intelligence (AI) researchers are increasingly concerned that whether the systems they build will be trusted by humans. Automated planning is one of the sub fields of AI that focuses on developing intelligent techniques to determine efficient plans, i.e., a sequence of actions that should be performed in order to achieve a set of goals.
Explainable AI Planning (XAIP) is a field that involves explaining AI planning systems to a user. The main goal of plan explanation is to help humans understand the plans produced by the planners. Approaches to this problem include explaining planner decision making processes as well as forming explanations from the models. Previous work on model-based explanations includes an iterative approach~\cite{smith12} as well as using explanations for more intuitive communication with the user~\cite{fox2017explainable}.

 Argumentation is connected to the idea of establishing trusted AI by explaining the results and processes of computation, and has been used in many applications in multi-agent planning~\cite{Torre_o_2018} and practical reasoning~\cite{AtkinsonB07}. This work is an attempt to generate explanation arguments in the domain of AI planning, to answer questions such as ‘Why A?’, where A is an action in the plan, or ‘How G?’, where G is a goal. Questions like these are inherently based upon definitions held in the domain related to a particular problem and solution. Furthermore, questions regarding particular state information may arise, such as ‘Why A here?’. Thus, extracting relevant information about actions, states and goals from the planning model is required to provide explanations to the user. Furthermore, some users might be interested in a summarized explanation of the whole plan and consequently inquire further information regarding the elements of the plan.

In this work, we make a first attempt to formalise a set of argument schemes~\cite{walton1996argumentation} that are aimed at creating arguments that explain and justify the plan and its key elements (i.e., action, state and goal). Furthermore, we present critical questions that allow the user to seek further information regarding the plan, and allow interaction between different arguments. Thus, the explanation arguments will enable a planning system to answer any such questions at a different granularity level.
To make our argumentation-based explanations for the planning study concrete, we take a version of the classic blocks world, as a case study example. 
% The rest of this paper is organised as follows. In section~\ref{sec:related-work}, we present the related work. In section~\ref{sec:planning-background}, we describe the planning model. In section~\ref{sec:argument-schemes}, we present the argumentation schemes for the explanation arguments, set of critical questions that allow argument interaction, and some plan argument properties. Finally, we present conclusions and future work in section~\ref{sec:conclusions}.

\section{Related Work}
\label{sec:related-work}

Our research is inspired by the works in practical reasoning and argumentation for multi-agent planning. However, our argument scheme based approach, generates explanations for a plan created by an AI planner, which we assume to be a single entity. One of the most well known scheme-based approach in practical reasoning is presented in~\cite{AtkinsonB07}, which is accompanied by a set of critical questions that allow agents to evaluate the outcomes on the basis of the social values highlighted by the arguments. Furthermore, in~\cite{TonioloNS11}, a model for arguments is presented that contributes in deliberative dialogues based on argumentation schemes for arguing about norms and actions in a multi-agent system. \cite{Oren13} has proposed a similar scheme-based approach for normative practical reasoning where arguments are constructed for a sequence of actions. 

\cite{ShamsVOP16} propose a framework that integrates both the reasoning and dialectical aspects of argumentation to perform normative practical reasoning, enabling an agent to act in a normative environment under conflicting goals and norms and generate explanation for agent behaviour. 
\cite{BelesiotisRR10} have explored the use of situation calculus as a language to present arguments about a common plan in a multi-agent system. \cite{TangP05} present an argumentation-based  approach  to deliberation, the process by which two or more agents reach a consensus on a course of action.

The works that are closest to our research for generating plan explanations using argumentation are given in~\cite{Caminada2014ScrutablePE} and~\cite{fan18}. In~\cite{Caminada2014ScrutablePE}, a dialectical proof based on the grounded semantics~\cite{CaminadaP12} is created to justify the actions executed in a plan. More recently, in~\cite{fan18}, an Assumption-based argumentation framework (ABA)~\cite{aba_Dung2009} is used to model the planning problem and generate explanation using the related admissible semantic~\cite{aaai-FanT15}. Our work differs from both, as we present argument schemes to generate the arguments that directly provide an explanation. Moreover, we use the concept of critical questions to provide dialectical interaction with the user and arguments. 
% Similar to~~\cite{Caminada2014ScrutablePE}, we use a planner to generate the plan and argumentation to generate explanations, differing from the later~\cite{fan18}, where argumentation is used to generate the plan and its explanation.

\section{Planning Model}
\label{sec:planning-background}

In this section, we introduce a planning model which is based on an instance of the most widely used planning representation, PDDL (Planning Domain Definition Language), as given in~\cite{book-pddl}. We define the planning problem as follows.

\begin{definition}{(Planning Problem)}
A planning problem is a tuple $P = \langle O, \mathit{Pr}, \bigtriangleup_I, \bigtriangleup_G, A, \Sigma, G \rangle$, where:
\setlist{nolistsep}
\begin{enumerate}[noitemsep]
    \item $O$ is a set of objects;
    \item $\mathit{Pr}$ is a set of predicates;
    \item $\bigtriangleup_I \subseteq \mathit{Pr}$ is the initial state;
    \item $\bigtriangleup_G \subseteq \mathit{Pr}$ is the goal state, and $G$ is the set of goals;   
    \item $A$ is a finite, non-empty set of actions;
    \item $\Sigma$ is the state transition system;    
\end{enumerate}
\end{definition}

\noindent We define the predicates as follows.

\begin{definition}{(Predicates)}
$\mathit{Pr}$ is a set of domain predicates, i.e., properties of objects that we are interested in, that can be true or false. For a state $s \subseteq Pr$, $s^+$ are predicates considered \textit{true}, and $s^- = Pr\setminus s^+$. A state $s$ satisfies predicate $pr$, denoted as $s \models pr$, if $pr \in s$, and satisfies predicate $\neg pr$, denoted $s \models \neg pr$, if $pr \not\in s$.
\end{definition}

\noindent We define two types of actions, the standard sequential action, i.e., action, and the concurrent action.

\begin{definition}{(Action)}
An action $a = \langle pre, post\rangle$ is composed of sets of predicates $pre$, $post$ that represent $a$'s pre and post conditions respectively. Given an action $a=\langle pre, post\rangle$, we write $pre(a)$ and $post(a)$ for $pre$ and $post$.
Postconditions are divided into $add(post(a)^+)$ and $delete(post(a)^-)$ postcondition sets. An action $a$ can be executed in state $s$ iff the state satisfies its preconditions. The postconditions of an action are applied in the state $s$ at which the action ends, by adding the positive postconditions belonging to $post(a)^+$ and deleting the negative postconditions belonging to $post(a)^-$.
\end{definition}

\begin{definition}{(Concurrent Action)}
A concurrent action $a_c$ is an action that can be concurrently executed with other concurrent actions. Two concurrent actions $a_i$ and $a_j$ (where $i \neq j$) are executable if their preconditions hold and their effects, i.e., postconditions are consistent. Furthermore, the effects of $a_i$ should not contradict the preconditions of $a_j$ and vice-versa.
\end{definition}

\noindent We define the state transition system as follows.

\begin{definition}{(State Transition System)}
The state-transition system is denoted by $\Sigma =(S,A,\gamma)$, where:
\setlist{nolistsep}
\begin{itemize}[noitemsep]
    \item $S$ is the set of states.
    \item $A$ is a finite, non-empty set of actions.
    \item $\gamma: S \times A = S$ where:
        \begin{itemize}[noitemsep]
            \item $\gamma(S,a) = (S \setminus post(a)^-)) \cup post(a)^+$, if $a$ is applicable in $S$;
            \item $\gamma(S,a) = \mathit{undefined}$ otherwise;
            \item $S$ is closed under $\gamma$.
        \end{itemize}
\end{itemize}
\end{definition}

\noindent We define the goal in a plan as follows.

\begin{definition}{(Goal)}
A goal achieves a certain state of affairs. Each $g \in G$ is a set of predicates $g=\{r_1,...,r_n\}$, known as \textit{goal requirements} (denoted as $r_i$), that should be satisfied in the state to satisfy the goal.
\end{definition}

\noindent We define a plan as follows.

\begin{definition}{(Plan)}
A plan is a sequence of actions $\pi = \langle a_1,...,a_n \rangle$. 
The extended state transition function for a plan is defined as follows:
\setlist{nolistsep}
\begin{itemize}[noitemsep]
    \item $\gamma(S, \pi) = S$ if $|\pi|=0$ (i.e., if $\pi$ is empty);
    \item $\gamma(S, \pi)= \gamma(\gamma(S,a_1),a_2,...,a_n)$ if $|\pi|>0$ and $a_1$ is applicable in $S$;
    \item $\gamma(S, \pi)= \mathit{undefined}$ otherwise.
\end{itemize}

A plan $\pi$ is a solution to a planning problem $P$ iff:
\setlist{nolistsep}
\begin{enumerate}[noitemsep]
    \item Only the predicates in $\bigtriangleup_I$ hold in the initial state: $S_1 = \bigtriangleup_I$;
    \item the preconditions of action $a_i$ hold at state $S_i$, where $i=1,2,...,n$;
    \item $\gamma(S_i,\pi)$ satisfies the set of goals $G$.
    \item the set of goals satisfied by plan $\pi$ is a non-empty $G_\pi \neq \emptyset$ consistent subset of goals.
\end{enumerate}
\end{definition}

Each action in the plan can be performed in the state that results from the application of the previous action in the sequence. After performing the final action, the set of goals $G_\pi$ will be true.
We use the following Blocks World example to illustrate.

\begin{example}
\label{example:blocks-world}

A classic blocks world consists of the following:
\begin{inparaenum}[(1)]
     \item A flat surface such as a tabletop,
    \item An adequate set of identical blocks which are identified by letters,
    \item The blocks can be stacked one on one to form towers of unlimited height.    
\end{inparaenum}

We have three predicates:
\setlist{nolistsep}
\begin{enumerate}[noitemsep]
    \item $\mathit{ON(A,B)}$ -- block $A$ is on block $B$. 
    \item $\mathit{ONTABLE(A)}$ -- block $A$ is on the table. 
    \item $\mathit{CLEAR(A)}$ -- block $A$ has nothing on it. 
\end{enumerate}

Following are the two actions $a_1$ and $a_2$:

\begin{enumerate}[noitemsep]
    \item $a_1: \mathit{UNSTACK(A,B)}$ -- pick up clear block $A$ from block $B$; 
        \begin{itemize}[noitemsep]
            \item $\mathit{pre(a_1)}: \mathit{CLEAR(A)} \wedge \mathit{ON(A,B)}$
            \item $\mathit{post(a_1)^+}: \mathit{ONTABLE(A)} \wedge \mathit{CLEAR(B)}$
            \item $\mathit{post(a_1)^-}: \mathit{ON(A,B)}$            
        \end{itemize}
    \item $a_2: \mathit{STACK(A,B)}$ -- place block $A$ onto clear block $B$; 
        \begin{itemize}[noitemsep]
            \item $\mathit{pre(a_2)}: \mathit{ONTABLE(A)} \wedge \mathit{CLEAR(A)} \wedge \mathit{CLEAR(B)}$
            \item $\mathit{post(a_2)^+}: \mathit{ON(A,B)}$
            \item $\mathit{post(a_2)^-}: \mathit{ONTABLE(A)} \wedge \mathit{CLEAR(B)}$            
        \end{itemize}    
\end{enumerate}

We have the following conditional statements:
\begin{itemize}[noitemsep]
    \item If block A is on the table it is not on any other block.
    \item If block A is on block B, block B is not clear. 
\end{itemize}

\noindent The initial and goal states of the blocks world problem are shown in Figure~\ref{fig:blocks_world}.
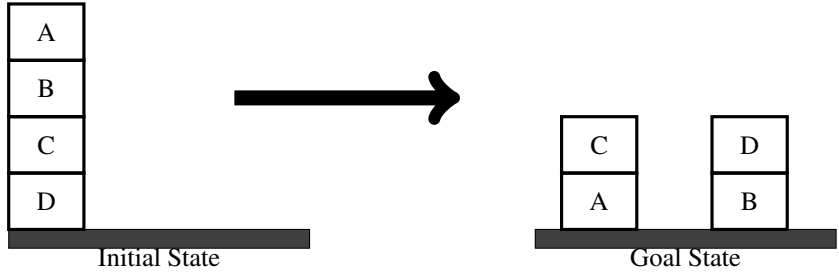
\begin{figure}[!h]
	\centering
\begin{tikzpicture}
 
\filldraw[draw=black,fill=darkgray] (0,0) rectangle node [below] {Initial State} (4,0.25);
% \node [text,align=center] [below=D] {Initial Goal};
%\node[box,label=below:stack] {Initial Goal};
\draw[black, very thick] (0,0.25) rectangle node{D} (1,1.00);
\draw[black, very thick] (0,1.00) rectangle node{C} (1,1.75);
\draw[black, very thick] (0,1.75) rectangle node{B} (1,2.50);
\draw[black, very thick] (0,2.50) rectangle node{A} (1,3.25);

\draw[->, line width=2mm] (3,2) -- (6,2);

\filldraw[draw=black,fill=darkgray] (7,0) rectangle node [below] {Goal State} (11,0.25);
%\node [text,align=center] [below=A] {Final Goal};
\draw[black, very thick] (7.35,0.25) rectangle node{A} (8.35,1.00);
\draw[black, very thick] (7.35,1.00) rectangle node{C} (8.35,1.75);

\draw[black, very thick] (9.35,0.25) rectangle node{B} (10.35,1.00);
\draw[black, very thick] (9.35,1.00) rectangle node{D} (10.35,1.75);
%\draw[orange, ultra thick] (4,0) -- (6,0) -- (5.7,2) -- cycle;
\end{tikzpicture}
	\caption{Blocks World Example}
	\label{fig:blocks_world}
\end{figure}

% \hfill\break
\noindent The initial state $\bigtriangleup_I$ is given by:

\noindent $\mathit{ONTABLE(D)} \wedge \mathit{ON(C,D)} \wedge \mathit{ON(B,C)} \wedge \mathit{ON(A,B)} \wedge \mathit{CLEAR(A)}$.

% \hfill\break
\noindent The goal state $\bigtriangleup_G$ is given by:

\noindent $\mathit{ON(C,A)} \wedge \mathit{ON(D,B)} \wedge \mathit{ONTABLE(A)} \wedge \mathit{ONTABLE(B)} \wedge \mathit{CLEAR(C)} \wedge \mathit{CLEAR(D)}$.

% \hfill\break
\noindent The action sequence: 

\noindent $\langle \mathit{UNSTACK(A,B)}, \mathit{UNSTACK(B,C)}, \mathit{UNSTACK(C,D)}, \mathit{(STACK(C,A), STACK(D,B))} \rangle$ is a solution plan.
\end{example}

\section{Argument Schemes for Explaining Plans}
\label{sec:argument-schemes}

In scheme-based approaches~\cite{walton1996argumentation} arguments are expressed in natural language and a set of critical questions is associated with each scheme, identifying how the scheme can be attacked. Below, we introduce a set of argument schemes for explaining a plan and its key elements, i.e., action, concurrent action, state and goal. The set of critical questions allow the user to ask for a summary explanation for the plan and consequently interrogate the elements of the plan. The explanation arguments constructed using the argument schemes allow the planner to answer any user questions.

\begin{definition}
Given a planning problem $P$:
\setlist{nolistsep}
\begin{itemize}[noitemsep]
    \item $\mathit{Hold(pre(a),S)}$ indicates that the precondition $pre(a)$ of action $a$ holds at state $S$.
    \item $\mathit{Execute(a,S)}$ indicates that action $a$ is executed at state $S$.
    \item $\mathit{ExecuteC(a_c,S)}$ indicates that all the concurrent actions in the set $a_c = \{a_0,a_1,...,a_n\}$ are executed at state $S$.    
    \item $\mathit{Achieve(a,g)}$ indicates that action $a$ achieves goals $g$.
    \item $\mathit{Solution(\pi, P)}$ indicates that $\pi$ is a solution to the planning problem $P$.
\end{itemize}
\end{definition}

\begin{definition}{(Action Argument Scheme $\mathit{Arg}_a$)}
An action argument $\mathit{Arg}_a$ explains how it is possible to execute an action $a$:
\setlist{nolistsep}
\begin{itemize}[noitemsep]
    \item \textbf{Premise $1$:} $\mathit{Hold(pre(a), S_1)}$. In the current state $S_1$, the pre-condition $pre(a)$ of action $a$ holds.
    \item \textbf{Premise $2$:} $\gamma(S_1, a) = S_2$. When we execute action $a$ in the current state $S_1$, it results in the next state $S_2$.
    \item  \textbf{Premise $3$}: $\mathit{Hold(g, S_2)}$. In the next state $S_2$, the goal $g$ holds.
    \item  \textbf{Premise $4$}. $\mathit{Achieve(a,g)}$: Action $a$ achieves goal $g$.
    \item  \textbf{Conclusion}: $\mathit{Execute(a, S_1)}$. Therefore, we should execute action $a$ in the current state $S_1$.    
\end{itemize}
\end{definition}

\begin{example}
\label{example:action-as}

We consider the blocks world of Example~\ref{example:blocks-world}. The explanation argument for the first action $\mathit{UNSTACK(A,B)}$ is shown as follows. Where:
\setlist{nolistsep}
\begin{itemize}[noitemsep]
    \item $pre(\mathit{UNSTACK(A,B)}) = \mathit{CLEAR(A) \wedge ON(A,B)}$.
    \item $S_1 = \mathit{ONTABLE(D) \wedge ON(C,D) \wedge ON(B,C) \wedge ON(A,B) \wedge CLEAR(A)}$.
    \item $S_2 = \mathit{ONTABLE(D) \wedge ON(C,D) \wedge ON(B,C) \wedge CLEAR(A) \wedge ONTABLE(A) \wedge CLEAR(B)}$.
    \item $g = \mathit{ONTABLE(A)}$.
\end{itemize}

     \noindent \textbf{Premise $1$:} 
     
     \noindent $\mathit{Hold(CLEAR(A) \wedge ON(A,B), \; ONTABLE(D) \wedge ON(C,D) \wedge ON(B,C) \wedge ON(A,B)} \wedge CLEAR(A))$
     
     \smallskip
     \noindent In the current state $\mathit{ONTABLE(D) \wedge ON(C,D) \wedge ON(B,C) \wedge ON(A,B) \wedge CLEAR(A)}$, the pre-condition $\mathit{CLEAR(A) \wedge ON(A,B)}$ of action $\mathit{UNSTACK(A,B)}$ holds.
     
     \smallskip
      \noindent \textbf{Premise $2$:} 
      
      \noindent $\mathit{\gamma(ONTABLE(D) \wedge ON(C,D) \wedge ON(B,C) \wedge ON(A,B) \wedge CLEAR(A), \; UNSTACK(A,B))}$
      
        \noindent $= \mathit{ONTABLE(D) \wedge ON(C,D) \wedge ON(B,C) \wedge CLEAR(A) \wedge ONTABLE(A) \wedge CLEAR(B)}$.
      
        % \noindent $s.t. \; \gamma (ONTABLE(D) \wedge ON(C,D) \wedge ON(B,C) \wedge ON(A,B),$
        
        % \noindent $\mathit{ONTABLE(D) \wedge ON(C,D) \wedge ON(B,C) \wedge CLEAR(A) \wedge ONTABLE(A) \wedge CLEAR(B)})$
      
      \smallskip
      \noindent When we execute action $\mathit{UNSTACK(A,B)}$ in the current state $\mathit{ONTABLE(D) \wedge ON(C,D) \wedge}$ 
      
      \noindent $\mathit{ON(B,C) \wedge ON(A,B) \wedge CLEAR(A)}$, it results in the next state $\mathit{ONTABLE(D) \wedge ON(C,D) \wedge}$ 
      
      \noindent $\mathit{ ON(B,C) \wedge CLEAR(A) \wedge ONTABLE(A) \wedge CLEAR(B)}$.
    %   , such that the state transition function 
      
    %   \noindent $\gamma(\mathit{ONTABLE(D) \wedge ON(C,D) \wedge ON(B,C) \wedge ON(A,B)}, \; \mathit{ONTABLE(D) \wedge ON(C,D) \wedge ON(B,C)}$ 
      
    %   \noindent $\mathit{\wedge CLEAR(A) \wedge ONTABLE(A) \wedge CLEAR(B)})$ holds.

    % \item $S_2 \gets \mathit{Excecute(a, S_1)}$: When we execute action $A$ in current state $S_1$, it results in next state $S_2$.
    \smallskip
     \noindent \textbf{Premise $3$:} 
     
    \noindent $\mathit{Hold(ONTABLE(A), \; ONTABLE(D) \wedge ON(C,D) \wedge ON(B,C) \wedge CLEAR(A) \wedge ONTABLE(A)}$
    
    \noindent $\mathit{\wedge CLEAR(B))}$
    
    \smallskip
    \noindent In the next state $\mathit{ONTABLE(D) \wedge ON(C,D) \wedge ON(B,C) \wedge CLEAR(A) \wedge ONTABLE(A) \wedge}$
    
    \noindent $\mathit{CLEAR(B)}$, the goal $\mathit{ONTABLE(A)}$ holds.

   \smallskip    
    \noindent \textbf{Premise $4$:} 
         
    \noindent$\mathit{Achieve(UNSTACK(A,B), \; ONTABLE(A))}$
    
    \smallskip
    \noindent Action $\mathit{UNSTACK(A,B)}$ achieves goal $\mathit{ONTABLE(A)}$.
    
    \smallskip
    \noindent \textbf{Conclusion:} 
    
    \noindent $\mathit{Excecute(UNSTACK(A,B), \; ONTABLE(D) \wedge ON(C,D) \wedge ON(B,C) \wedge}$
    
    \noindent $\mathit{ON(A,B) \wedge CLEAR(A))}$    
    
    \smallskip
    \noindent Therefore, we should execute action $\mathit{UNSTACK(A,B)}$ in the current state $\mathit{ONTABLE(D) \wedge}$ 
    $\mathit{ON(C,D) \wedge ON(B,C) \wedge ON(A,B) \wedge CLEAR(A)}$.    

% Precondition of action $a_0$ is given by $\mathit{pre(a_0)}: \mathit{CLEAR(A)} \wedge \mathit{ON(A,B)}$. The current state (which is also the initial state) is given by $S_1: \mathit{ONTABLE(D)} \wedge \mathit{ON(C,D)} \wedge \mathit{ON(B,C)} \wedge \mathit{ON(A,B)} \wedge \mathit{CLEAR(A)}$. Thus, the $pre(a_0)$ holds in state $S_1$. 

\end{example}

\begin{definition}{(Concurrent Action Argument Scheme $\mathit{Arg}_{a_c}$)}
A concurrent action argument $\mathit{Arg}_{a_c}$ explains how it is possible to execute all concurrent actions in the set $a_c = \{a_1, a_2,...,a_n\}$. 
% Two concurrent actions $a_i$ and $a_j$ (where $i \neq j$) are executable if their preconditions hold and their effects are consistent. Furthermore, the effects of $a_i$ should not contradict the preconditions of $a_j$ and vice-versa.
\setlist{nolistsep}
\begin{itemize}[noitemsep]
    \item \textbf{Premise $1$:} $\mathit{Hold(pre(a_1), S_1)} \wedge \mathit{Hold(pre(a_2), S_1)} \wedge ... \wedge \mathit{Hold(pre(a_n), S_1)}$. In the current state $S_1$, the preconditions of all the concurrent actions in the set $a_c$ hold.
    \item \textbf{Premise $2$:} $\forall a_i,a_j \in a_c \: (\mathit{where} \: i \neq j) \: \gamma(S_1,a_i) = S_2 \; \wedge$
    
    \noindent $\mathit{Hold(pre(a_j),S_2)}$. When we execute any concurrent action $a_i$ in the state $S_1$, it results in the state $S_2$, and the precondition $pre(a_j)$ of any other concurrent action $a_j$ holds in the state $S_2$.
    \item \textbf{Premise $3$:} $\gamma(S_n,a_n) = S_G$. When we execute the last concurrent action $a_n$ in the state $S_n$, it results in the next state $S_G$.
    \item \textbf{Premise $4$:} $\mathit{Hold(G,S_G)}$. In the next state $S_G$, the set of goals $G$ holds.
    \item \textbf{Premise $5$:} $\mathit{Achieve(a_c,G)}$. The set of concurrent actions $a_c$ achieves the set of goals $G$.
    \item \textbf{Conclusion:} $\mathit{ExecuteC(a_c,S_1)}$. Therefore, we should execute all the concurrent actions in the set $a_c$ in the current state $S_1$.    
    % \item In the current state $S_1$.
    % \item The preconditions of all the concurrent actions in the set $A_c$ hold.
    % \item Execution of any action $a \in a_c$ results in a new state $S_2$ where preconditions of all actions in the set $a_c \setminus a$ hold.
    % \item Therefore, we should execute all concurrent actions in the set $a_c$.
    % \item Which will result in new state $S_G$.
    % \item Which will realise a set of goals $G$.
\end{itemize}
\end{definition}

\begin{example}
\label{example:caction-as}

The concurrent action argument $\mathit{Arg}_{a_c}$ for the set of concurrent actions $ a_c = \{  \mathit{STACK(C,A), \; STACK(D,B)} \}$ in the Example~\ref{example:blocks-world} is shown as follows. Where:
\setlist{nolistsep}
\begin{itemize}[noitemsep]
\item $\mathit{pre(STACK(C,A))} = \mathit{ONTABLE(C)} \wedge \mathit{CLEAR(C)} \wedge \mathit{CLEAR(A)}$, 
\item $\mathit{pre(STACK(D,B))} = \mathit{ONTABLE(D)} \wedge \mathit{CLEAR(D)} \wedge \mathit{CLEAR(B)}$, 
\item $S_1 = \mathit{ONTABLE(D)} \wedge \mathit{CLEAR(A)} \wedge \mathit{CLEAR(B)} \wedge \mathit{CLEAR(C)} \wedge \mathit{CLEAR(D)} \wedge$

\noindent $\mathit{ONTABLE(A)} \wedge \mathit{ONTABLE(B)} \wedge \mathit{ONTABLE(C)}$.
\item $S_2 = \mathit{ONTABLE(D) \wedge CLEAR(B) \wedge CLEAR(C) \wedge CLEAR(D) \wedge ON(C,A)  \wedge ONTABLE(A)}$ 

\noindent $\mathit{\wedge ONTABLE(B)}$
\item $S_G = \mathit{CLEAR(C) \wedge CLEAR(D) \wedge ON(C,A) \wedge ON(D,B) \wedge ONTABLE(A)  \wedge ONTABLE(B)}$.

\item $G = \{ \mathit{ON(C,A)}, \mathit{ON(D,B)} \}$, i.e., the set of all the goals in the goal state $S_G$ that are not in the state $S_1$.

\end{itemize}

% $S_1 = \mathit{ONTABLE(D)} \wedge \mathit{CLEAR(A)} \wedge \mathit{CLEAR(B)} \wedge \mathit{CLEAR(C)} \wedge \mathit{CLEAR(D)} \wedge \mathit{ONTABLE(A)} \wedge \mathit{ONTABLE(B)} \wedge \mathit{ONTABLE(C)}$.

    %\smallskip
     \noindent \textbf{Premise $1$:} 
     
     \noindent $\mathit{Hold}(\mathit{ONTABLE(C)} \wedge \mathit{CLEAR(C)} \wedge \mathit{CLEAR(A)}, \; \mathit{ONTABLE(D)} \wedge \mathit{CLEAR(A)} \wedge \mathit{CLEAR(B)} \wedge$
     
     \noindent $\mathit{CLEAR(C)} \wedge \mathit{CLEAR(D)} \wedge \mathit{ONTABLE(A)} \wedge \mathit{ONTABLE(B)} \wedge \mathit{ONTABLE(C)})$
     
     \noindent $\wedge \; \mathit{Hold}(\mathit{ONTABLE(D)} \wedge \mathit{CLEAR(D)} \wedge \mathit{CLEAR(B)}, \;
     \mathit{ONTABLE(D)} \wedge \mathit{CLEAR(A)} \wedge \mathit{CLEAR(B)} \wedge \mathit{CLEAR(C)} \wedge \mathit{CLEAR(D)} \wedge \mathit{ONTABLE(A)} \wedge \mathit{ONTABLE(B)} \wedge \mathit{ONTABLE(C)})$.
     
     \smallskip
     \noindent In the current state $\mathit{ONTABLE(D)} \wedge \mathit{CLEAR(A)} \wedge \mathit{CLEAR(B)} \wedge \mathit{CLEAR(C)} \wedge \mathit{CLEAR(D)} \wedge \mathit{ONTABLE(A)} \wedge \mathit{ONTABLE(B)} \wedge \mathit{ONTABLE(C)}$, the precondition $\mathit{ONTABLE(C)} \wedge \mathit{CLEAR(C)} \wedge \mathit{CLEAR(A)}$ of action $STACK(C,A)$ holds and the precondition $\mathit{ONTABLE(D)} \wedge \mathit{CLEAR(D)} \wedge \mathit{CLEAR(B)}$ of action $STACK(D,B)$ holds.
     
     \smallskip
     \noindent \textbf{Premise $2$:} 
     
     \noindent $(\gamma(\mathit{ONTABLE(D)} \wedge \mathit{CLEAR(A)} \wedge \mathit{CLEAR(B)} \wedge \mathit{CLEAR(C)} \wedge \mathit{CLEAR(D)} \wedge \mathit{ONTABLE(A)} \wedge \mathit{ONTABLE(B)} \wedge \mathit{ONTABLE(C)}, \; \mathit{STACK(C,A)})$
     
     \noindent $= \mathit{ONTABLE(D) \wedge CLEAR(B) \wedge CLEAR(C) \wedge CLEAR(D) \wedge ON(C,A)  \wedge ONTABLE(A)}$
     
     \noindent $\wedge \mathit{ONTABLE(B)})\;  \wedge \; \mathit{Hold}(\mathit{ONTABLE(D)} \wedge \mathit{CLEAR(D)} \wedge \mathit{CLEAR(B)}, \; \mathit{ONTABLE(D) \wedge CLEAR(B)}$
     
     \noindent $\mathit{\wedge CLEAR(C) \wedge CLEAR(D) \wedge ON(C,A)  \wedge ONTABLE(A) \wedge ONTABLE(B)})$.
     
     \smallskip
     \noindent When we execute the concurrent action $STACK(C,A)$ in the state $\mathit{ONTABLE(D)} \wedge \mathit{CLEAR(A)} \wedge \mathit{CLEAR(B)} \wedge \mathit{CLEAR(C)} \wedge \mathit{CLEAR(D)} \wedge \mathit{ONTABLE(A)} \wedge \mathit{ONTABLE(B)} \wedge \mathit{ONTABLE(C)}$, it results in the next state $\mathit{ONTABLE(D) \wedge CLEAR(B) \wedge CLEAR(C) \wedge CLEAR(D)}$
     
     \noindent $\mathit{\wedge ON(C,A)  \wedge ONTABLE(A) \wedge ONTABLE(B)}$, and the precondition $\mathit{ONTABLE(D)} \wedge \mathit{CLEAR(D)} \wedge \mathit{CLEAR(B)}$ of the other concurrent action $STACK(D,B)$ holds in the  next state $\mathit{ONTABLE(D) \wedge CLEAR(B) \wedge CLEAR(C) \wedge CLEAR(D) \wedge ON(C,A)  \wedge ONTABLE(A) \wedge}$ 
     
     \noindent $\mathit{ONTABLE(B)}$.
     
     \smallskip
     \noindent \textbf{Premise $3$:} 
     
     \noindent$\gamma(\mathit{ONTABLE(D) \wedge CLEAR(B) \wedge CLEAR(C) \wedge CLEAR(D) \wedge ON(C,A)  \wedge ONTABLE(A) \wedge}$
     
     \noindent $\mathit{ONTABLE(B)}, \; STACK(D,B)) = \mathit{CLEAR(C) \wedge CLEAR(D) \wedge ON(C,A) \wedge ON(D,B) \wedge ONTABLE(A)}$
     
     \noindent $\mathit{\wedge ONTABLE(B)}$.
     
    %  \; s.t. \; \gamma(\mathit{ONTABLE(D) \wedge CLEAR(B) \wedge CLEAR(C) \wedge CLEAR(D) \wedge ON(C,A)  \wedge}$
     
    %  \noindent $\mathit{ONTABLE(A)}, \;  \mathit{CLEAR(C) \wedge CLEAR(D) \wedge ON(C,A) \wedge ON(D,B) \wedge ONTABLE(A)  \wedge}$
     
    %  \noindent$\mathit{ONTABLE(B)})$. 
     
     \smallskip
     \noindent When we execute the last concurrent action $\mathit{STACK(D,B)}$ in the state $\mathit{ONTABLE(D) \wedge CLEAR(B)}$
     
     \noindent $\mathit{\wedge CLEAR(C) \wedge CLEAR(D) \wedge ON(C,A)  \wedge ONTABLE(A) \wedge ONTABLE(B)}$, it results in the next state $\mathit{CLEAR(C) \wedge CLEAR(D) \wedge ON(C,A) \wedge ON(D,B) \wedge ONTABLE(A)  \wedge ONTABLE(B)}$.
     
     \smallskip
     \noindent \textbf{Premise $4$:} 
     
     \noindent $\mathit{Hold}(\{ \mathit{ON(C,A)}, \mathit{ON(D,B)} \}, \; \mathit{CLEAR(C) \wedge CLEAR(D) \wedge ON(C,A) \wedge ON(D,B) \wedge ONTABLE(A)}$  
     
     \noindent $\mathit{\wedge ONTABLE(B)})$. 
     
     \smallskip
     \noindent In the next state $\mathit{CLEAR(C) \wedge CLEAR(D) \wedge ON(C,A) \wedge ON(D,B) \wedge ONTABLE(A)  \wedge ONTABLE(B)}$, the set of goals $\{ \mathit{ON(C,A)}, \mathit{ON(D,B)} \}$ holds.
     
    %\newpage
    \smallskip
    \noindent \textbf{Premise $5$:} 
    
    \noindent $\mathit{Achieve}( \{  \mathit{STACK(C,A), STACK(D,B)} \}, \; \{ \mathit{ON(C,A)}, \mathit{ON(D,B)} \} )$. 
    
    \smallskip
    \noindent The set of concurrent actions $ \{  \mathit{STACK(C,A), STACK(D,B)} \}$ achieves the set of goals $\{ \mathit{ON(C,A)}, \mathit{ON(D,B)} \}$.
     
    \smallskip
    \noindent \textbf{Conclusion:} 
    
    \noindent $\mathit{ExecuteC}(\{  \mathit{STACK(C,A), STACK(D,B)} \}, \; \mathit{ONTABLE(D)} \wedge \mathit{CLEAR(A)} \wedge \mathit{CLEAR(B)} \wedge \mathit{CLEAR(C)} \wedge \mathit{CLEAR(D)} \wedge \mathit{ONTABLE(A)} \wedge \mathit{ONTABLE(B)} \wedge \mathit{ONTABLE(C)} )$. 
    
    \smallskip
    \noindent Therefore, we should execute all the concurrent actions in the set $\{  \mathit{STACK(C,A), STACK(D,B)} \}$ in the current state $\mathit{ONTABLE(D)} \wedge \mathit{CLEAR(A)} \wedge \mathit{CLEAR(B)} \wedge \mathit{CLEAR(C)} \wedge \mathit{CLEAR(D)} \wedge \mathit{ONTABLE(A)} \wedge \mathit{ONTABLE(B)} \wedge \mathit{ONTABLE(C)}$.    

\end{example}

\begin{definition}{(State Transition Argument Scheme $\mathit{Arg}_S$)}
A state transition argument $\mathit{Arg}_S$ explains how the state $S$ becomes true:
\setlist{nolistsep}
\begin{itemize}[noitemsep]
    \item \textbf{Premise $1$:} 
    $\gamma(S_1,a) = (S_1 - post(a)^-)) \cup post(a)^+ = S$.
    % $\gamma(S_1,a) = S$. 
    In the current state $S_1$, we should execute the action $a \in \pi$ by deleting the negative postconditions $post(a)^-$ and adding the positive postconditions $post(a)^+$ to $S_1$, that results in the state $S$.
    \item \textbf{Conclusion:} Therefore, the state $S$ is true.
\end{itemize}
\end{definition}

\begin{example}
\label{example:state-as}

The state transition argument $\mathit{Arg}_S$ for the state $ S = \mathit{ONTABLE(D)} \wedge \mathit{ON(C,D)} \wedge \mathit{ON(B,C)} \wedge \mathit{CLEAR(A)} \wedge \mathit{CLEAR(B)} \wedge \mathit{ONTABLE(A)}$ in the Example~\ref{example:blocks-world} is shown as follows. Where:
\setlist{nolistsep}
\begin{itemize}[noitemsep]
    \item $a = \mathit{UNSTACK(A,B)}$.
    \item $post(a)^- = \mathit{ON(A,B)}$
    \item $post(a)^+ = \mathit{ONTABLE(A) \wedge CLEAR(B)}$
    \item $S_1 = \mathit{ONTABLE(D)} \wedge \mathit{ON(C,D)} \wedge \mathit{ON(B,C)} \wedge \mathit{ON(A,B)} \wedge \mathit{CLEAR(A)}$.
\end{itemize}

    %\smallskip
     \noindent \textbf{Premise $1$:} 
     
     \noindent $\gamma(\mathit{ONTABLE(D)} \wedge \mathit{ON(C,D)} \wedge \mathit{ON(B,C)} \wedge \mathit{ON(A,B)} \wedge \mathit{CLEAR(A)},\; \mathit{UNSTACK(A,B)})$
     
     \noindent $= (\mathit{ONTABLE(D)} \wedge \mathit{ON(C,D)} \wedge \mathit{ON(B,C)} \wedge \mathit{ON(A,B)} \wedge \mathit{CLEAR(A)} - \mathit{ON(A,B)}) \cup \mathit{ONTABLE(A) \wedge CLEAR(B)}$
     
     \noindent $= \mathit{ONTABLE(D)} \wedge \mathit{ON(C,D)} \wedge \mathit{ON(B,C)} \wedge \mathit{CLEAR(A)} \wedge \mathit{CLEAR(B)} \wedge \mathit{ONTABLE(A)}$.
     
     \smallskip
     \noindent In the current state $\mathit{ONTABLE(D)} \wedge \mathit{ON(C,D)} \wedge \mathit{ON(B,C)} \wedge \mathit{ON(A,B)} \wedge \mathit{CLEAR(A)}$, we should execute the action $\mathit{UNSTACK(A,B)}$ by deleting the negative postconditions $\mathit{ON(A,B)}$ and adding the positive postconditions $\mathit{ONTABLE(A) \wedge CLEAR(B)}$ to the current state $\mathit{ONTABLE(D)} \wedge \mathit{ON(C,D)} \wedge \mathit{ON(B,C)} \wedge \mathit{ON(A,B)} \wedge \mathit{CLEAR(A)}$, that results in the state $\mathit{ONTABLE(D)} \wedge \mathit{ON(C,D)} \wedge \mathit{ON(B,C)} \wedge \mathit{CLEAR(A)} \wedge \mathit{CLEAR(B)} \wedge \mathit{ONTABLE(A)}$.
     
     \smallskip
     \noindent \textbf{Conclusion:} 
     \noindent Therefore, the state $\mathit{ONTABLE(D)} \wedge \mathit{ON(C,D)} \wedge \mathit{ON(B,C)} \wedge \mathit{CLEAR(A)} \wedge \mathit{CLEAR(B)} \wedge \mathit{ONTABLE(A)}$ is true.

\end{example}

\begin{definition}{(Goal Argument Scheme $\mathit{Arg}_g$)}
A goal argument $\mathit{Arg}_g$ explains how a \textit{feasible goal}\footnote{A goal is feasible if there is at least one plan that satisfies it.} is achieved by an action in the plan:
\setlist{nolistsep}
\begin{itemize}[noitemsep]
    \item \textbf{Premise $1$:} $\gamma(S_1,a) = S_2$. In the current state $S_1$, we should execute the action $a \in \pi$, that results in the next state $S_2$.
    \item \textbf{Premise $2$:} $\mathit{Hold(g, S_2)}$. In the next state $S_2$, the goal $g$ holds.
    \item \textbf{Conclusion:} $\mathit{Achieve(a,g)}$: Therefore, the action $a$ achieves the goal $g$.    
    % \item In the initial state $\bigtriangleup$.
    % \item Sequence of actions $\delta \subseteq \pi$ is executed.
    % \item Which will result in the new state $S_r$.
    % \item The goals is satisfied in the resulting state $S_r$.
    % \item Therefore, goal $g$ is achievable.
\end{itemize}
% The set of goal argument is denoted as $\mathit{Arg}_G$.
\end{definition}

\begin{example}
\label{example:goal-as}

The goal argument $\mathit{Arg}_g$ for the goal $g = \mathit{ONTABLE(A)}$ in the Example~\ref{example:blocks-world} is shown as follows. Where:
\setlist{nolistsep}
\begin{itemize}[noitemsep]
    \item $a = \mathit{UNSTACK(A,B)}$.
    \item $S_1 = \mathit{ONTABLE(D)} \wedge \mathit{ON(C,D)} \wedge \mathit{ON(B,C)} \wedge \mathit{ON(A,B)} \wedge \mathit{CLEAR(A)}$.
    \item $S_2 = \mathit{ONTABLE(D)} \wedge \mathit{ON(C,D)} \wedge \mathit{ON(B,C)} \wedge \mathit{CLEAR(A)} \wedge \mathit{CLEAR(B)} \wedge \mathit{ONTABLE(A)}$.
\end{itemize}

    %\smallskip
     \noindent \textbf{Premise $1$:} 
    %  [td, ocd, obc, na, nb, ta]
    %  $\mathit{ONTABLE(D)} \wedge \mathit{ON(C,D)} \wedge \mathit{ON(B,C)} \wedge \mathit{CLEAR(A)} \wedge \mathit{CLEAR(B)} \wedge \mathit{ONTABLE(A)}$

     \noindent $\gamma(\mathit{ONTABLE(D)} \wedge \mathit{ON(C,D)} \wedge \mathit{ON(B,C)} \wedge \mathit{ON(A,B)} \wedge \mathit{CLEAR(A)}, \; \mathit{UNSTACK(A,B)})$
     
     \noindent $= \mathit{ONTABLE(D)} \wedge \mathit{ON(C,D)} \wedge \mathit{ON(B,C)} \wedge \mathit{CLEAR(A)} \wedge \mathit{CLEAR(B)} \wedge \mathit{ONTABLE(A)}$.
     
     \smallskip
     \noindent In the current state $\mathit{ONTABLE(D)} \wedge \mathit{ON(C,D)} \wedge \mathit{ON(B,C)} \wedge \mathit{ON(A,B)} \wedge \mathit{CLEAR(A)}$, we should execute the action $\mathit{UNSTACK(A,B)}$, that results in the next state $\mathit{ONTABLE(D)} \wedge \mathit{ON(C,D)} \wedge \mathit{ON(B,C)} \wedge \mathit{CLEAR(A)} \wedge \mathit{CLEAR(B)} \wedge \mathit{ONTABLE(A)}$.
     
     \smallskip
     \noindent \textbf{Premise $2$:} 
     
     \noindent $\mathit{Hold}(\mathit{ONTABLE(A)}, \; \mathit{ONTABLE(D)} \wedge \mathit{ON(C,D)} \wedge \mathit{ON(B,C)} \wedge \mathit{CLEAR(A)} \wedge \mathit{CLEAR(B)} \wedge \mathit{ONTABLE(A)})$. 
     
     \smallskip
     \noindent In the next state $\mathit{ONTABLE(D)} \wedge \mathit{ON(C,D)} \wedge \mathit{ON(B,C)} \wedge \mathit{CLEAR(A)} \wedge \mathit{CLEAR(B)} \wedge \mathit{ONTABLE(A)}$, the goal $\mathit{ONTABLE(A)}$ holds.
     
          \smallskip
     \noindent \textbf{Conclusion:} 
     
     \noindent $\mathit{Achieve(\mathit{UNSTACK(A,B)}, \mathit{ONTABLE(A)})}$: 
     
     \smallskip
     \noindent Therefore, the action $\mathit{UNSTACK(A,B)}$ achieves the goal $\mathit{ONTABLE(A)}$. 

\end{example}

\begin{definition}{(Plan Summary Argument Scheme $\mathit{Arg}_{\pi}$)}
A plan summary argument $\mathit{Arg}_{\pi}$ explains that a proposed sequence of actions $\pi=\langle a_1,a_2,...,a_n \rangle$ is a solution to the planning problem $P$ because it achieves a set of goals $G$:
\setlist{nolistsep}
\begin{itemize}[noitemsep]
    % \item \textbf{Premise $1$:} $\gamma(\bigtriangleup_G, \mathit{Execute(\pi, \bigtriangleup_I)})$. In the initial state $\bigtriangleup_I$, we should execute the sequence of actions $\pi$, that result in the goal state $\bigtriangleup_G$.
    \item \textbf{Premise $1$:} 
    % $\gamma(\gamma(\bigtriangleup_I,a_1),a_2,...,a_n) = \bigtriangleup_G$.
    $\gamma(S_1,a_1) = S_2$, $\gamma(S_2,a_2) = S_3$,...,$\gamma(S_n,a_n) = S_{n+1}$.
    % $S_n \gets 
    % \mathit{Execute(a_n, \; \bigtriangleup_{n-1} \gets Execute(a_{n-1}, \; \bigtriangleup_{n-2}...Execute(a_1, \; S_1) ) )}$ s.t. $\gamma(\bigtriangleup_G, \bigtriangleup_I)$ where $S_n = \bigtriangleup_G$ and $S_1 = \bigtriangleup_I$. 
    In the initial state $S_1 = \bigtriangleup_I$, we should execute the first action $a_1$ in the sequence of actions $\pi$ that results in the next state $S_2$ and execute the next action $a_2$ in the sequence in the state $S_2$ that results in the next state $S_3$ and carry on until the last action $a_n$ in the sequence is executed in the state $S_n$ that results in the goal state $S_{n+1}=\bigtriangleup_G$.
    \item \textbf{Premise $2$:} $\mathit{Hold(G, \bigtriangleup_G)}$. In the goal state $\bigtriangleup_G$, all the goals in the set of goals $G$ hold.
    \item \textbf{Premise $3$:} $\mathit{Achieve(\pi,G)}$. The sequence of actions $\pi$ achieves the set of all goals $G$.      
    \item \textbf{Conclusion:} $\mathit{Solution(\pi, P)}$. Therefore, $\pi$ is a solution to the planning problem $P$.
\end{itemize}
\end{definition}

\begin{example}
\label{example:plan-as}

The plan summary argument $\mathit{Arg}_{\pi}$ for the solution plan given in the Example~\ref{example:blocks-world} is shown as follows. 
    
    \smallskip
     \noindent \textbf{Premise $1$:} 
     
%      \noindent $\gamma (\mathit{ON(C,A)} \wedge \mathit{ON(D,B)} \wedge \mathit{ONTABLE(A)} \wedge \mathit{ONTABLE(B)} \wedge \mathit{CLEAR(C)} \wedge \mathit{CLEAR(D)},$

% \noindent $\mathit{Execute}(\mathit{UNSTACK(A,B)}, \mathit{UNSTACK(B,C)}, \mathit{UNSTACK(C,D)}, \mathit{(STACK(C,A), STACK(D,B)),}$ 

% \noindent $\mathit{ONTABLE(D)} \wedge \mathit{ON(C,D)} \wedge \mathit{ON(B,C)} \wedge \mathit{ON(A,B)} \wedge \mathit{CLEAR(A)})$.

    \noindent $\gamma(\mathit{ONTABLE(D) \wedge ON(C,D) \wedge ON(B,C) \wedge ON(A,B) \wedge CLEAR(A)}, \; \mathit{UNSTACK(A,B)})$
    
    \noindent $ = \mathit{ONTABLE(D) \wedge ON(C,D) \wedge ON(B,C) \wedge CLEAR(A) \wedge CLEAR(B) \wedge ONTABLE(A)}$,
    
    \smallskip
    \noindent $\gamma(\mathit{ONTABLE(D) \wedge ON(C,D) \wedge ON(B,C) \wedge CLEAR(A) \wedge CLEAR(B) \wedge ONTABLE(A)},$
    
    \noindent $\mathit{UNSTACK(B,C)}) = \mathit{ONTABLE(D) \wedge ON(C,D) \wedge CLEAR(A) \wedge CLEAR(B) \wedge CLEAR(C) \wedge}$
    
    \noindent $\mathit{ONTABLE(A) \wedge ONTABLE(B)}$,
    
    \smallskip
    \noindent $\gamma(\mathit{ONTABLE(D) \wedge ON(C,D) \wedge CLEAR(A) \wedge CLEAR(B) \wedge CLEAR(C) \wedge ONTABLE(A) \wedge}$
    
    \noindent $\mathit{ONTABLE(B)}, \; \mathit{UNSTACK(C,D)}) = \mathit{ONTABLE(D) \wedge CLEAR(A) \wedge CLEAR(B) \wedge CLEAR(C)}$
    
    \noindent $\mathit{\wedge CLEAR(D) \wedge ONTABLE(A) \wedge ONTABLE(B) \wedge ONTABLE(C)}$,
    
    \smallskip
    \noindent $\gamma(\mathit{ONTABLE(D) \wedge CLEAR(A) \wedge CLEAR(B) \wedge CLEAR(C) \wedge CLEAR(D) \wedge ONTABLE(A) \wedge}$ 
    
    \noindent $\mathit{ONTABLE(B) \wedge ONTABLE(C)}, \; (STACK(C,A), STACK(D,B))) = \mathit{ON(C,A) \wedge ON(D,B) \wedge}$
    \noindent $\mathit{ONTABLE(A) \wedge ONTABLE(B) \wedge CLEAR(C) \wedge CLEAR(D)} $
     
     \smallskip

     \noindent In the initial state $\mathit{ONTABLE(D)} \wedge \mathit{ON(C,D)} \wedge \mathit{ON(B,C)} \wedge \mathit{ON(A,B)} \wedge \mathit{CLEAR(A)}$, we should execute the action $\mathit{UNSTACK(A,B)}$ that results in the next state $\mathit{ONTABLE(D) \wedge}$
     
     \noindent $\mathit{ON(C,D) \wedge ON(B,C) \wedge CLEAR(A) \wedge CLEAR(B) \wedge ONTABLE(A)}$.
     
     \smallskip
     \noindent In the state $\mathit{ONTABLE(D) \wedge ON(C,D) \wedge ON(B,C) \wedge CLEAR(A) \wedge CLEAR(B) \wedge ONTABLE(A)}$, we should execute the action $\mathit{UNSTACK(B,C)}$ that results in the next state $\mathit{ONTABLE(D) \wedge}$
     
     \noindent $\mathit{ON(C,D) \wedge CLEAR(A) \wedge CLEAR(B) \wedge CLEAR(C) \wedge ONTABLE(A) \wedge ONTABLE(B)}$.

     \smallskip
     \noindent In the state $\mathit{ONTABLE(D) \wedge ON(C,D) \wedge CLEAR(A) \wedge CLEAR(B) \wedge CLEAR(C) \wedge ONTABLE(A)}$ 
     
     \noindent $\mathit{\wedge ONTABLE(B)}$, we should execute the action $\mathit{UNSTACK(C,D)}$ that results in the next state $\mathit{ONTABLE(D) \wedge CLEAR(A) \wedge CLEAR(B) \wedge CLEAR(C) \wedge CLEAR(D) \wedge ONTABLE(A) \wedge}$
     
     \noindent $\mathit{ONTABLE(B) \wedge ONTABLE(C)}$.
     
     \smallskip
     \noindent In the state $\mathit{ONTABLE(D) \wedge CLEAR(A) \wedge CLEAR(B) \wedge CLEAR(C) \wedge CLEAR(D) \wedge ONTABLE(A)}$
     
     \noindent $\mathit{\wedge ONTABLE(B) \wedge ONTABLE(C)}$, we should execute all the concurrent actions in the set $\mathit{(STACK(C,A), STACK(D,B))}$ that result in the goal state $\mathit{ON(C,A) \wedge ON(D,B) \wedge ONTABLE(A)}$
     
     \noindent $\mathit{\wedge ONTABLE(B) \wedge CLEAR(C) \wedge CLEAR(D)}$.
    %  \noindent In the initial state $\mathit{ONTABLE(D)} \wedge \mathit{ON(C,D)} \wedge \mathit{ON(B,C)} \wedge \mathit{ON(A,B)} \wedge \mathit{CLEAR(A)}$, we should execute the sequence of actions $\mathit{UNSTACK(A,B)}, \mathit{UNSTACK(B,C)}, \mathit{UNSTACK(C,D)},$ 
    %  \noindent $\mathit{(STACK(C,A), STACK(D,B))}$, that result in the goal state $\mathit{ON(C,A)} \wedge \mathit{ON(D,B)} \wedge \mathit{ONTABLE(A)} \wedge \mathit{ONTABLE(B)} \wedge \mathit{CLEAR(C)} \wedge \mathit{CLEAR(D)}$.
     
     \smallskip
      \noindent \textbf{Premise $2$:} 
      
      \noindent $ \mathit{Hold}(\{ \mathit{ON(C,A)}, \mathit{ON(D,B)}, \mathit{ONTABLE(A)}, \mathit{ONTABLE(B)}, \mathit{CLEAR(C)}, \mathit{CLEAR(D)} \},$ 
      
      \noindent $\mathit{ON(C,A)} \wedge \mathit{ON(D,B)} \wedge \mathit{ONTABLE(A)} \wedge \mathit{ONTABLE(B)} \wedge \mathit{CLEAR(C)} \wedge \mathit{CLEAR(D)})$.
      
      \smallskip
      \noindent In the goal state $\mathit{ON(C,A)} \wedge \mathit{ON(D,B)} \wedge \mathit{ONTABLE(A)} \wedge \mathit{ONTABLE(B)} \wedge \mathit{CLEAR(C)} \wedge \mathit{CLEAR(D)}$, all the goals in the set of goals $\{ \mathit{ON(C,A)}, \mathit{ON(D,B)}, \mathit{ONTABLE(A)}, \mathit{ONTABLE(B)},$
      
      \noindent $\mathit{CLEAR(C)}, \mathit{CLEAR(D)} \}$ hold.
      
     \smallskip
      \noindent \textbf{Premise $3$:}      
      
      \noindent $\mathit{Achieve}(\langle \mathit{UNSTACK(A,B)}, \mathit{UNSTACK(B,C)}, \mathit{UNSTACK(C,D)}, \mathit{(STACK(C,A), STACK(D,B))} \rangle,$ 
      
      \noindent $ \{ \mathit{ON(C,A)}, \mathit{ON(D,B)}, \mathit{ONTABLE(A)}, \mathit{ONTABLE(B)}, \mathit{CLEAR(C)}, \mathit{CLEAR(D)} \})$.
      
      \smallskip
      \noindent The sequence of actions $\langle \mathit{UNSTACK(A,B)}, \mathit{UNSTACK(B,C)}, \mathit{UNSTACK(C,D)}, \mathit{(STACK(C,A),}$ 
      
      \noindent $\mathit{STACK(D,B))} \rangle$ achieves the set of all goals $ \{ \mathit{ON(C,A)}, \mathit{ON(D,B)}, \mathit{ONTABLE(A)}, \mathit{ONTABLE(B)},$
      
      \noindent $\mathit{CLEAR(C)}, \mathit{CLEAR(D)} \}$.    
      
     \smallskip
      \noindent \textbf{Conclusion:}   
      
      \noindent $\mathit{Solution(\langle \mathit{UNSTACK(A,B)}, \mathit{UNSTACK(B,C)}, \mathit{UNSTACK(C,D)}, \mathit{(STACK(C,A), STACK(D,B))} \rangle, P)}$. 
      
      \smallskip
      \noindent Therefore, $\langle \mathit{UNSTACK(A,B)}, \mathit{UNSTACK(B,C)}, \mathit{UNSTACK(C,D)}, \mathit{(STACK(C,A), STACK(D,B))} \rangle$ is a solution to the planning problem $P$.

\end{example}

\subsection{Argument Interactions and Properties of Plan Arguments}

The five critical questions (CQs) given below describe the ways in which the arguments built using the argument schemes can interact with each other. These CQs are associated to (i.e., attack) one or more premises of the arguments constructed using the argument schemes and are in turn answered (i.e., attacked) by the other arguments, which are listed in the description.

\smallskip
\noindent \textbf{CQ1: Is the plan $\pi$ possible?} This CQ begins the dialogue with the user, and it is the first question that the user asks when presented with a solution plan $\pi$. The argument scheme $\mathit{Arg}_{\pi}$, answers the CQ by constructing the summary argument for the plan $\pi$.

\smallskip
\noindent \textbf{CQ2: Is the action $a$ possible to execute?} This CQ is associated with the following argument schemes: $\mathit{Arg}_{\pi}$, $\mathit{Arg}_{S}$, $\mathit{Arg}_{g}$. The argument scheme $\mathit{Arg}_{a}$, answers the CQ by constructing the explanation argument for the action $a$.

\smallskip
\noindent \textbf{CQ3: Is the set of concurrent actions $a_c$ possible to execute?} This CQ is associated with the argument schemes $\mathit{Arg}_{\pi}$. The argument scheme $\mathit{Arg}_{a_c}$, answers the CQ by constructing the explanation argument for the set of concurrent actions $a_c$.

\smallskip
\noindent \textbf{CQ4: Is the state $S$ possible?} This CQ is associated with the following argument schemes: $\mathit{Arg}_{\pi}$, $\mathit{Arg}_{a}$, $\mathit{Arg}_{a_c}$, $\mathit{Arg}_{g}$. The argument scheme $\mathit{Arg}_{S}$, answers the CQ by constructing the explanation argument for the state $S$.

\smallskip
\noindent \textbf{CQ5: Is the goal $g$ possible to achieve?} This CQ is associated with the argument scheme $\mathit{Arg}_{\pi}$. The argument scheme $\mathit{Arg}_{g}$, answers the CQ by constructing the explanation argument for the goal $g$.

We organise the arguments and their interactions, as presented earlier, by mapping into a Dung $\mathit{AF} = (\mathcal{A}, \mathcal{R})$~\cite{Dung95onthe}, where $\mathcal{A}$ is a set of arguments and $\mathcal{R}$ is an attack relation $(\mathcal{R} \subseteq \mathcal{A} \times \mathcal{A})$. $\mathit{Args} \subset \mathcal{A}$ and $\mathit{CQs} \subset \mathcal{A}$, where $\mathit{Args} = \{\mathit{Arg}_{\pi}, \mathit{Arg}_a, \mathit{Arg}_{c_a}, \mathit{Arg}_S, \mathit{Arg}_g\}$ and $\mathit{CQs} = \{\mathit{CQ}_1, \mathit{CQ}_2, \mathit{CQ}_3, \mathit{CQ}_4, \mathit{CQ}_5\}$. We use the  grounded  extension  of  the  $\mathit{AF}$, denoted by $\mathit{Gr}$, to determine if a plan should be acceptable. 

\smallskip
\noindent \textbf{Property 1:}
For a plan $\pi$, the set of arguments $\mathit{Args}$ is complete, in that, if a $\mathit{CQ} \in \mathit{CQs}$ exists, then it will be answered (i.e., attacked) by an $\mathit{Arg} \in \mathit{Args}$.

\noindent \textit{Proof.} Since, $(\mathit{Arg}_{\pi}, CQ1) \in \mathcal{R}$, $(\mathit{Arg}_a, CQ2) \in \mathcal{R}$, $(\mathit{Arg}_{a_c}, CQ3) \in \mathcal{R}$, $(\mathit{Arg}_S, CQ4) \in \mathcal{R}$, and $(\mathit{Arg}_g, CQ5) \in \mathcal{R}$, therefore, a unique $\mathit{Arg} \in \mathit{Args}$ exists, that attacks a unique $\mathit{CQ} \in \mathit{CQs}$. Thus, $\mathit{Args}$ is complete.

\smallskip
\noindent \textbf{Property 2:}
For a plan $\pi$, $\mathit{Arg}_{\pi} \in \mathit{Gr}$ iff $CQ \not\in \mathit{Gr}$ when $(\mathit{CQ}, \mathit{Arg}_{\pi}) \in \mathcal{R}$, $CQ \in CQs$.

\noindent \textit{Proof.} Follows from Property 1. Since any CQ that attacks $\mathit{Arg}_{\pi}$ is in turn attacked by an $\mathit{Arg} \in \mathit{Args}$, therefore, $CQ \not\in \mathit{Gr}$. Thus, $\mathit{Arg}_{\pi} \in \mathit{Gr}$.

\smallskip
\noindent \textbf{Property 3:}
For a plan $\pi$, $\mathit{Arg}_{\pi} \in \mathit{Gr}$ iff $\forall g \in G \; \mathit{Arg}_g \in \mathit{Gr}$.

\noindent \textit{Proof.} Since plan $\pi$ achieves all goals $g \in G$, and $\mathit{Arg}_{g\in G}$ attack all CQs that attack the goal premises of $\mathit{Arg}_{\pi}$, therefore, $\forall g \in G \; \mathit{Arg}_g \in \mathit{Gr}$. Thus, $\mathit{Arg}_{\pi} \in \mathit{Gr}$.

\smallskip
\noindent \textbf{Property 4:} For a plan $\pi$, $\mathit{Arg}_{\pi} \in \mathit{Gr}$ iff the explanation is acceptable to the user.

\noindent \textit{Proof.} Follows immediately from Properties 1, 2 and 3.

\section{Conclusions and Future Work}
\label{sec:conclusions}
In this paper, we have presented a novel argument scheme-based approach for creating explanation arguments in the domain of AI planning. The main contributions of our work are as follows:
\begin{inparaenum}[(i)]
    \item We formalise a set of argument schemes that help in constructing arguments that explain a plan and its key elements;
    \item We present critical questions that allow the user to seek further information regarding the key elements of the plan, and the interaction between different arguments;
    \item We present properties of the plan arguments.
\end{inparaenum}

In the future, we aim to develop algorithms based on the argument schemes to automatically extract the arguments from the input planning model. Furthermore, we intend to build a scheme-based dialogue system for creating interactive dialectical explanations. Finally, our approach to generating explanation arguments is planner independent and therefore, can work on a wide range of input plans in classical planning, and in the future, we intend to extend this to richer planning formalisms such as  partial order planning.

\smallskip
\noindent \textbf{Acknowledgments:} This  work  was supported by EPSRC grant EP/R033722/1, Trust in Human-Machine Partnership.

 \bibliographystyle{abbrv}
\bibliography{refs}
\end{document}